# Adaptive Distribution-aware Quantization for Mixed-Precision Neural Networks

Shaohang Jia, Zhiyong Huang*, Zhi Yu, Mingyang Hou, Shuai Miao, Han Yang

*School of Microelectronics and Communication Engineering, Chongqing University, Chongqing 400044, China*

\*Corresponding author: zyhuang@cqu.edu.cn (Z. Huang)

Abstract

Quantization-Aware Training (QAT) is a critical technique for deploying deep neural networks on resource-constrained devices. However, existing methods often face two major challenges: the highly non-uniform distribution of activations and the static, mismatched codebooks used in weight quantization. To address these challenges, we propose Adaptive Distribution-aware Quantization (ADQ), a mixed-precision quantization framework that employs a differentiated strategy. The core of ADQ is a novel adaptive weight quantization scheme comprising three key innovations: (1) a quantile-based initialization method that constructs a codebook closely aligned with the initial weight distribution; (2) an online codebook adaptation mechanism based on Exponential Moving Average (EMA) to dynamically track distributional shifts; and (3) a sensitivity-informed strategy for mixed-precision allocation. For activations, we integrate a hardware-friendly non-uniform-to-uniform mapping scheme. Comprehensive experiments validate the effectiveness of our method. On ImageNet, ADQ enables a ResNet-18 to achieve 71.512% Top-1 accuracy with an average bit-width of only 2.81 bits, outperforming state-of-the-art methods under comparable conditions. Furthermore, detailed ablation studies on CIFAR-10 systematically demonstrate the individual contributions of each innovative component, validating the rationale and effectiveness of our design.

## 1 Introduction

Deep Convolutional Neural Networks (CNNs) have become the cornerstone of modern computer vision, achieving remarkable success across numerous tasks. However, their high computational and memory requirements pose significant challenges for deployment on resource-constrained platforms such as mobile phones and edge devices. To overcome these hurdles, researchers have explored various model compression techniques, including network pruning to remove redundant parameters [19], knowledge distillation to transfer knowledge from large teacher models to smaller student models [16], and lightweight architecture design. Among these, model quantization, which represents weights and activations with low-bit integers, is particularly prominent. Unlike methods that alter the network architecture, quantization directly replaces resource-intensive floating-point operations with efficient integer arithmetic, offering the most direct path to inference acceleration on modern hardware and making it an indispensable technology for practical deployment.

Quantization-Aware Training (QAT) [1], which simulates the effects of quantization during training, typically yields much higher accuracy than Post-Training Quantization (PTQ). Despite its success, the efficacy of QAT is often constrained by a "one-size-fits-all" approach that neglects the distinct statistical properties of weights and activations. On one hand, the quantization of activations in CNNs is exceptionally challenging. As shown in Figure 1.1, their distributions are often non-uniform, exhibiting long tails or severe skewness, and they vary dynamically across different layers and even channels. Early methods like DoReFa-Net [2], which apply a fixed uniform quantizer, lead to significant information loss. Although subsequent approaches such as PACT [3] and LSQ [4] introduce learnable parameters to adjust the range of uniform quantization, they still struggle to effectively model the underlying non-uniformity.

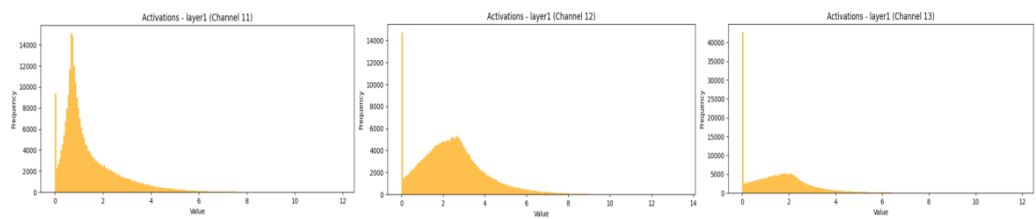

Figure 1.1 Histogram of activation values for channels 11–13 in the first layer of ResNet20

On the other hand, weights typically follow a more stable, bell-shaped distribution, as depicted in Figure 1.2. Yet, they present their own set of challenges. Uniform quantization wastes valuable encoding space in low-density regions, while existing non-uniform methods (e.g., LQ-Nets [5]) often rely on static, pre-defined codebooks that fail to adapt to the subtle shifts in weight distributions during training, resulting in suboptimal performance. Furthermore, the assumption that all layers share a uniform sensitivity to quantization is invalid. Different layers exhibit varying degrees of robustness to quantization noise, which has motivated the adoption of mixed-precision quantization techniques. While sophisticated methods based on Hessian analysis [6, 14, 15] or Neural Architecture Search (NAS) [7] exist to automate bit-width allocation, they often introduce substantial computational overhead.

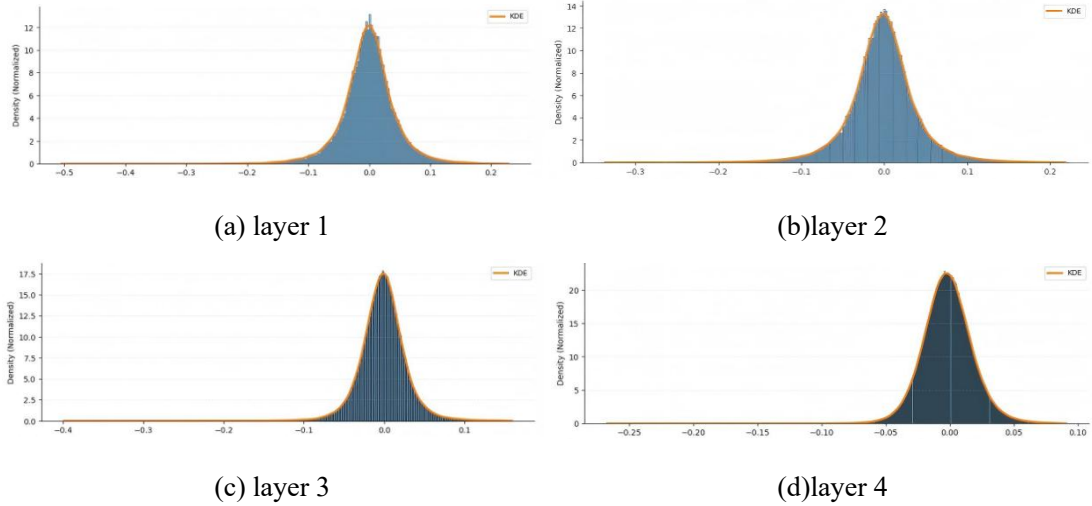

(a) layer 1　　　　　　　　　　　　(b) layer 2

(c) layer 3　　　　　　　　　　　　(d) layer 4

Figure 1.2 Weight Distribution of ResNet18 Layers 1–4

To tackle these multifaceted challenges, we introduce Adaptive Distribution-aware Quantization (ADQ), a novel framework built on a "divide and conquer" philosophy. We recognize that no single quantization strategy is optimal for both weights and activations. For activations, we employ a hardware-friendly non-uniform-to-uniform mapping scheme, whose efficacy has been validated in prior work [8], to effectively preserve information without compromising inference efficiency.

The primary contribution of this paper is a new, fully adaptive quantization scheme designed specifically for weights. This scheme features three main innovations:

- **Quantile-based Initialization:** We devise a method to initialize quantization codebooks directly from the quantiles of the pre-trained weight distribution, ensuring a near-optimal starting point for QAT.
- **Online Codebook Adaptation:** We employ an efficient Exponential Moving Average (EMA) mechanism that enables the codebook to dynamically track the evolution of the weight distribution during training.
- **Sensitivity-Aware Mixed-Precision:** We introduce a lightweight, heuristic-based strategy to assign different bit-widths to different layers, maximizing model accuracy under a given computational budget.

Comprehensive experiments demonstrate that ADQ sets a new state-of-the-art, validating the significant advantages of this adaptive and differentiated approach to quantization.

## 2 Related Work

### 2.1 Quantization Paradigms

Model quantization can be broadly categorized into Post-Training Quantization (PTQ) and Quantization-Aware Training (QAT). PTQ quantizes a pre-trained model using a small calibration dataset and is highly efficient. In contrast, QAT simulates the quantization process during training or fine-tuning, allowing the model to adapt to quantization errors and typically achieving higher accuracy. Our work, ADQ, falls under the QAT paradigm. A central challenge in QAT is backpropagating gradients through the non-differentiable quantization function. The Straight-Through Estimator (STE) [9] remains the most common solution, treating the quantizer as an identity function during the backward pass. Most advanced QAT methods, including ours, are built upon this principle.

### 2.2 Distribution-Aware Quantization

Given that weight and activation distributions are often non-uniform, many studies have moved beyond simple uniform quantization.

**For Activations:** The long-tailed nature of activation distributions makes them particularly sensitive to quantization. Many methods aim to adapt the quantization scheme to this characteristic. LSQ [4] and LSQ+ [10] achieve this by learning the scaling factor and zero-point of a uniform quantizer. Other approaches, such as DSQ [24] and QIL [25], introduce differentiable parameters to learn the quantization intervals themselves. More advanced non-uniform methods like LCQ [26] employ a learnable companding function to transform the data distribution before applying uniform quantization. The "non-uniform to uniform" mapping principle, as validated by N2UQ [8], is particularly effective and hardware-friendly, as it learns non-uniform input thresholds while generating uniform output levels.

**For Weights:** Although weights exhibit a more regular, bell-shaped distribution, applying a uniform grid is not optimal. Non-uniform methods like LQ-Nets [5] learn a codebook of quantization levels but typically assume this codebook is static. Recently, the concept of re-balancing has gained traction, especially for Large Language Models (LLMs). Methods such as SmoothQuant [11] and AWQ [12] use mathematical transformations to shift the quantization difficulty from activations with outliers to the weights. While effective, these methods add complexity. Our work addresses the weight distribution more directly by creating a codebook that is initialized based on the true distribution and dynamically adapts to it during training.

2.3 Mixed-Precision Quantization

Using a uniform bit-width for all layers is a rigid constraint, as different layers have varying sensitivities to quantization. Mixed-precision quantization addresses this by intelligently allocating bits. The core challenge is determining the optimal bit-width for each layer without an exhaustive search. Current approaches can be classified as follows:

**Search-Based Methods:** These methods frame bit allocation as a Neural Architecture Search (NAS) problem, using techniques like Differentiable NAS (DNAS) [7] or evolutionary algorithms [13] to find the optimal configuration. Differentiable Dynamic Quantization (DDQ) [22] also falls into this category, enabling dynamic bit-width adjustment during training. While effective, these methods are often associated with extremely high computational search costs.

**Sensitivity-Driven Methods:** A more practical approach is to approximate each layer's sensitivity to quantization using a metric. HAWQ and its variants [14, 15] pioneered this area, using the trace of the Hessian matrix as a robust sensitivity metric. However, even computing an approximation of the Hessian can be complex. Other works have proposed alternative metrics, but the goal remains the same: to find a reliable proxy for sensitivity. Our method is inspired by this line of research but proposes a more lightweight heuristic to balance accuracy and efficiency.

3 Method

3.1 Overall Framework

We propose Adaptive Distribution-aware Quantization (ADQ), a comprehensive mixed-precision QAT framework. As illustrated in Figure 3.1, ADQ processes weights and activations through two separate, specialized quantization paths. This "divide and conquer" strategy allows us to apply the most suitable quantization technique based on the unique statistical properties of each tensor type. For activations, we adopt a hardware-friendly scheme to handle their dynamic and non-uniform nature. For weights, we introduce a novel, fully adaptive quantization scheme, which constitutes the core contribution of this work.

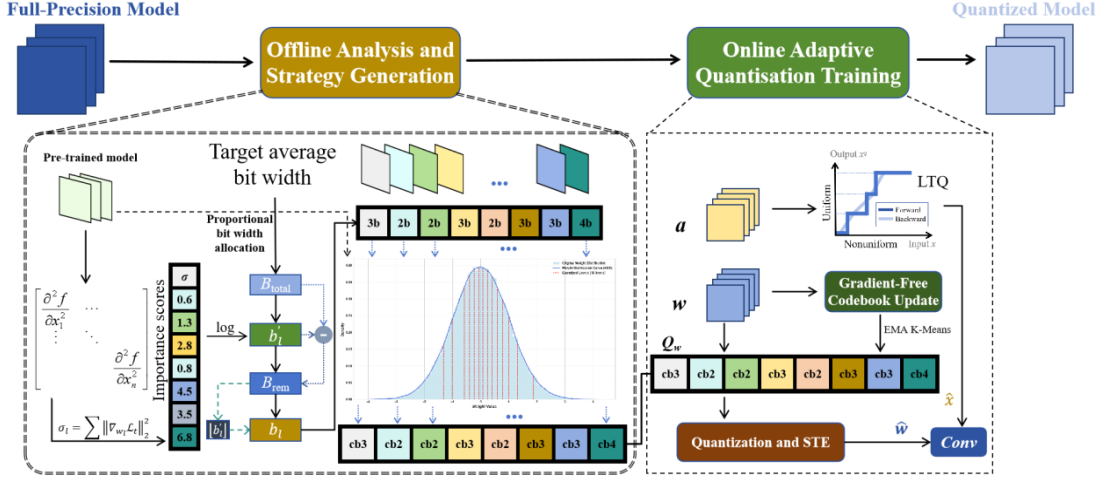

Figure 3.1 Overall framework of our proposed Adaptive Distribution-Aware Quantization (ADQ).

## 3.2 Hardware-Friendly Activation Quantization

To effectively quantize activations, which exhibit highly dynamic and skewed distributions, it is crucial to employ a strategy that adapts to their non-uniformity. An effective principle is to partition the input space non-uniformly, allocating higher resolution to high-density regions, while ensuring the output quantization levels remain uniform to maintain hardware efficiency. This "non-uniform-to-uniform" mapping has been shown to be highly effective in prior work [8]. Accordingly, our ADQ framework implements an activation quantizer based on this principle. Specifically, a set of learnable non-uniform thresholds $\{T_1, T_2, ..., T_{2^n-1}\}$ is used to partition the real-valued input into intervals, each mapping to a uniform output level. The forward pass for an n-bit quantizer is defined as:

$$x^q = \begin{cases} 0 & x^r < T_1 \\ 1 & T_1 \leq x^r < T_2 \\ \cdots & \cdots \\ 2^n - 1 & x^r \geq T_{2^n-1} \end{cases} \quad (1)$$

To enable the learning of these thresholds, gradients are approximated during backpropagation using a Generalized Straight-Through Estimator (G-STE) [9]. This allows the quantizer to learn an optimal data-driven partitioning scheme for each layer's activation distribution.

## 3.3 Adaptive Weight Quantization

The core of our ADQ framework is an innovative three-stage adaptive quantization scheme designed specifically for weights. This scheme draws inspiration from Vector Quantization (VQ) [17] and introduces mechanisms for dynamic, data-driven codebook generation and maintenance. It addresses key limitations of existing methods: poor initialization, static codebooks, and rigid bit allocation.

### 3.3.1 Quantile-based Codebook Initialization

**Motivation:** Random or uniform codebook initialization is severely mismatched with the typical bell-shaped weight distributions found in pre-trained models. This initial discrepancy can lead to large quantization errors and destabilize the initial phase of QAT, thereby hindering convergence.

**Method:** We propose a data-driven initialization method that leverages the pre-trained full-precision model. For the weight tensor of each layer, we construct a tailored initial codebook that reflects its underlying distribution. As shown in Algorithm 1, the weight tensor is first flattened, and its positive and negative values are separated. We then compute the quantiles for each set to generate representative sets of positive and negative quantization levels. These levels, combined with zero, form a symmetric and distribution-aware initial codebook $C_0$. This design ensures that each interval of the cumulative distribution function (CDF) receives equivalent representation.

---

Algorithm 1: Quantile-based Codebook Initialization

**Input**: Pre−trained weights $W_{fp}$, bit−width $b$.
**Output**: Initial codebook $C_0$.
1: $N \leftarrow 2^b$
2: $N_{pos} \leftarrow \lfloor (N-1)/2 \rfloor$, $N_{neg} \leftarrow \lceil (N-1)/2 \rceil$
3: $W_{pos}, W_{neg} \leftarrow \{|w| \text{ for } w \in W_{fp} \text{ if } w > 0\}, \{|w| \text{ for } w \in W_{fp} \text{ if } w < 0\}$
4: $P_{pos} \leftarrow \text{torch.linspace}(0, 1, N_{pos} + 2)[1:-1]$
5: $Q_{pos} \leftarrow \text{torch.quantile}(W_{pos}, P_{pos})$
6: (Repeat for negative values to get $Q_{neg}$)
7: $C_0 \leftarrow \text{Sort}(\text{Combine}(Q_{pos}, -Q_{neg}, \{0\}))$
8: **return** $C_0$

---

This initialization ensures that from the very beginning of training, our codebook provides an excellent representation of the weight distribution, laying a solid foundation for subsequent fine-tuning.

### 3.3.2 Online Codebook Adaptation via EMA-based K-Means

**Motivation:** During QAT, the weight distribution is not static; it evolves as the network is fine-tuned. A fixed codebook, even if well-initialized, will eventually become suboptimal.

**Method:** We introduce a lightweight and efficient mechanism to dynamically adjust the codebook online, which can be viewed as an EMA-smoothed variant of the K-Means clustering algorithm. This process involves two synergistic components:

a. Weight Quantization and Commitment Loss: In the forward pass, we first

normalize the weights *w* of each layer by a per-channel scaling factor *s*. For each normalized weight *w'*, we find the nearest codebook entry from the codebook $C=\{c_1, c_2, ..., c_N\}$. This constitutes the quantization step:

$$\hat{w}' = C(w') = \mathrm{argmin}_{c_i \in C} \|w' - c_i\|_2^2 \qquad (2)$$

To encourage the original weights to evolve toward the learned codebook, we introduce a "commitment loss" $\mathcal{L}_{commit}$, inspired by VQ-VAE. This loss term penalizes the distance between the weights and their chosen codebook representations, with its gradient flowing back only to the weights:

$$\mathcal{L}_{commit} = \beta \cdot \|\mathrm{sg}(\hat{w}') - w'\|_2^2 \qquad (3)$$

where $\mathrm{sg}(\cdot)$ is the stop-gradient operator and $\beta$ is a hyperparameter. This loss is added to the main task loss during training. The final quantized weights for convolution are reconstructed via the Straight-Through Estimator (STE):

$$\hat{w} = s \cdot (w' + (\hat{w}' - w').detach()) \qquad (4)$$

    b. Gradient-Free Codebook Update: The codebook itself is not updated via standard backpropagation. Instead, we employ Exponential Moving Average (EMA) for in-place centroid updates. We maintain moving averages of the sum of weights assigned to each codebook entry, $E_i$, and the count of assignments, $n_i$. At each training step, the updates are:

$$n_i^{(t)} = \alpha \cdot n_i^{(t-1)} + (1 - \alpha) \cdot \sum_j \mathbb{I}(C(w_j') = c_i) \qquad (5)$$

$$E_i^{(t)} = \alpha \cdot E_i^{(t-1)} + (1 - \alpha) \cdot \sum_{j, C(w_j')=c_i} w_j' \qquad (6)$$

where α is the EMA decay factor. The codebook is then updated to the new centroids of these evolving clusters:

$$c_i^{(t+1)} = \frac{E_i^{(t)}}{n_i^{(t)} + \epsilon} \qquad (7)$$

This enables the codebook to smoothly track the weight clusters with negligible computational overhead.

3.3.3 Sensitivity-based Mixed-Precision Allocation

    **Motivation:** Uniform bit-width allocation is resource-inefficient. Our goal is to develop a practical method for mixed-precision allocation that avoids the high costs of NAS or full Hessian computation.

    **Method:** We propose a lightweight, three-step heuristic strategy, detailed in Algorithm 2.

    a. Sensitivity Scoring: We first estimate the sensitivity of each layer to quantization. The score for layer *l* is calculated as the sum of squared gradients of its weights, accumulated over a small number of training batches. This metric serves as a

practical proxy for the diagonal of the Fisher Information Matrix [18], which has been shown to be an effective measure of parameter importance [19].

$$\sigma_l = \sum_{i=1}^{N_{batches}} ||\nabla_{w_l}\mathcal{L}_i||_2^2 \tag{8}$$

b. Proportional Bit Allocation: Instead of simple ranking, we adopt a continuous bit-width allocation: the bits $b_l'$ assigned to each layer are proportional to the logarithm of its sensitivity score. This ensures that highly sensitive layers receive a significantly larger bit budget, while moderately scaling the differences between less sensitive layers.

$$b_l' = B_{total} \cdot \frac{\log(1+\sigma_l)}{\sum_k \log(1+\sigma_k)} \tag{9}$$

where $B_{total}$ is the total bit budget across all layers.

c. Greedy Discretization: The continuous bit-widths are then discretized to available integer values (e.g., {2, 3, 4}). We use a greedy algorithm that first rounds down each layer's bit-width to the nearest available integer, then iteratively increments the bit-width of the layer with the largest fractional part until the model's target average bit-width is met.

Algorithm 2: Sensitivity-based Mixed-Precision Allocation
**Input**: Sensitivity scores $\{\sigma_l\}$, target average bit−width $B_{avg}$, available bits $B_{set}$
**Output**: Final bit assignment $\{b_l\}$
1: Calculate total budget $B_{total} = L \cdot B_{avg}$
2: For each layer $l$, Calculate continuous bits $\{b_l'\}$ using Formula 9.
3: For each layer $l$, initialize $b_l = \max\{\{b \in B_{set} | b \leq b_l'\}\}$.
4: Calculate remaining budget $B_{rem} = \text{round}(B_{total} - \sum_l b_l)$.
5: Calculate upgrade priorities $p_l = b_l' - b_l$.
6: **for** $i = 1$ to $B_{rem}$ **do**
7: $l^* = \text{argmax}_l(p_l)$
8: $b_{l^*} \leftarrow \min(\{b \in B_{set} | b > b_{l^*}\})$
9: $p_{l^*} \leftarrow -\infty$ (to prevent re−selection)
10: **end for**
11: **return** $\{b_l\}$

This sensitivity-driven approach ensures that the limited bit budget is allocated where it is needed most, protecting the most critical layers from significant quantization error in a highly efficient manner.

## 4 Experiments
### 4.1 Experimental Setup

**Datasets and Models:** Our main experiments are conducted on the ImageNet (ILSVRC 2012) [20] dataset. We use ResNet-18 [21] as the backbone network for core comparisons and ablation studies.

**Implementation Details:** Our framework is implemented in PyTorch. All models are trained using an SGD optimizer with an initial learning rate of 0.3, which is decayed to zero using a cosine annealing schedule over 120 epochs. For experiments involving Knowledge Distillation (KD) [16], we use the same KD scheme as in N2UQ [8]. Following standard practice, the first convolutional layer and the final fully-connected layer are kept at full precision to maintain feature richness and classification accuracy. The remaining convolutional layers within the main ResNet blocks are quantized using our ADQ method. Unless otherwise specified, weights and activations of each layer are quantized to the same bit-width.

4.2 Main Results on ImageNet

To validate the effectiveness of ADQ, we compare it with several state-of-the-art quantization methods on the ImageNet dataset. As shown in Table 4.1, our method achieves superior performance.

Table 4.1 Comparison with state-of-the-art methods on ImageNet (ResNet-18)

| Network | Method | FP(%) | W/A Bits | KD | Top-1 Acc. (%) |
|---|---|---|---|---|---|
| ResNet-18 | PACT[3] | 70.4 | 3/3 | ✗ | 68.1 |
| | LQ-Net[5] | 70.3 | 3/3 | ✗ | 68.2 |
| | DSQ[24] | 69.9 | 3/3 | ✗ | 68.7 |
| | QIL[25] | 70.2 | 3/3 | ✗ | 69.2 |
| | LSQ+[10] | 70.1 | 3/3 | ✗ | 69.4 |
| | QuanDCL[27] | 69.8 | 3/3 | ✗ | 69.5 |
| | LSQ[4] | 70.5 | 3/3 | ✗ | 70.2 |
| | LCQ[26] | 70.4 | 3/3 | ✗ | 70.6 |
| | DDQ[22] | 70.5 | 4/4 | ✓ | 71.2 |
| | QKD[28] | 70.1 | 3/3 | ✓ | 70.2 |
| | N2UQ [8] | 71.8 | 3/3 | ✓ | 71.9 |
| | HAWQ-V3[23] | 71.5 | 4/4 | ✗ | 68.5 |
| | ADQ (Ours) | 70.9 | 2.81 | ✓ | 71.5 |
| | ADQ (Ours) | | 2.81 | ✗ | 69.5 |

With knowledge distillation, our ADQ model, using a mixed-precision configuration (averaging only 2.81 bits), achieves a Top-1 accuracy of 71.5%. This result is highly competitive, surpassing numerous methods that use a higher, uniform

3-bit quantization scheme. Critically, even without KD, our method still achieves 69.5% accuracy. This strong standalone performance demonstrates the intrinsic effectiveness of our adaptive weight quantization scheme and provides a fair baseline for comparison with other non-distillation methods.

### 4.3 Ablation Studies

To dissect the contributions of the different components within the ADQ framework, we conducted a series of ablation studies.

#### 4.3.1 Effect of Knowledge Distillation

Knowledge distillation is a powerful technique for boosting the performance of quantized models. To distinguish the gains from ADQ itself versus the contribution of KD, we trained a ResNet-18 model with a ResNet-101 teacher and one without. As shown in Table 4.2, using our quantization method alone achieves a strong accuracy of 69.5%. The introduction of KD provides a significant 2% boost, reaching the final accuracy of 71.5%. This confirms that ADQ is a powerful quantization method in its own right and its benefits are complementary to standard training enhancements like KD.

Table 4.2 Ablation analysis of knowledge distillation (ImageNet, ResNet-18)

| Network | Configuration | Top-1 Acc. (%) |
|---|---|---|
| | FP32 Baseline | 70.8 |
| ResNet-18 | ADQ (w/o KD) | 69.5 |
| | ADQ (with KD) | 71.5 |

#### 4.3.2 Mixed-Precision Assignment Analysis

Our ADQ framework employs a sensitivity-based strategy to assign different bit-widths to different layers. For the ResNet-18 experiment, this resulted in the mixed-precision configuration shown in Table 4.3, with an average bit-width of 2.81. The allocation strategy clearly prioritizes deeper, more sensitive layers with 3 bits, while assigning 2 bits to the less critical, earlier layers in the first stage. This intelligent allocation is key to balancing accuracy and compression.

Table 4.3 Mixed-precision bit allocation for ResNet-18.

| Layers | Conv Layers | Bit-width（W/A） |
|---|---|---|
| layer1.0 | conv1, conv2 | 3,2 |
| layer1.1 | conv1, conv2 | 2,2 |
| layer2.x | all | 3 |
| layer3.x | all | 3 |
| layer4.x | all | 3 |

### 4.3.3 Analysis of ADQ Core Components

To validate the effectiveness of the three core components of our ADQ method—codebook initialization, online learning, and mixed precision—we conducted a series of ablation experiments on the CIFAR-10 dataset using a ResNet-20 model. As shown in Table 4.4, we first reproduce N2UQ as a fixed 3-bit baseline (Config. A). Our full method is Config. E (mixed 2.8-bit), with an additional Config. F (mixed 3-bit) for a fair comparison. To isolate components, we designed Config. B (randomly initialized codebook), Config. C (static codebook), and Config. D (fixed 3-bit ADQ). Since the original N2UQ paper did not report results for ResNet-20 on CIFAR-10, we built a fair baseline for comparison based on their open-source code and the core ideas described in their paper. Specifically, we used their LTQ for activations and uniform weight quantization. To ensure a fair comparison, this baseline shared the exact same training hyperparameters as all our proposed method variants, including the optimizer, learning rate schedule, data augmentation, and number of training epochs.

Table 4.4: Ablation of ADQ components (CIFAR-10, ResNet-20)

| Config. | Codebook Init | Codebook Learned | Bit Allocation | Top-1 Acc. (%) |
|---|---|---|---|---|
| A. Baseline (N2UQ) | - | - | Fixed 3-bit | 88.21 |
| B. ADQ (Random Init) | $\mathcal{N}(0,1)$ | ✓ | Mixed 2.8-bit | 90.26 |
| C. ADQ (Static Codebook) | Based on pre-trained model | ✗ | Mixed 2.8-bit | 89.95 |
| D. ADQ (Fixed 3-bit) | Based on pre-trained model | ✓ | Fixed 3-bit | 90.98 |
| E. Full ADQ (Ours) | Based on pre-trained model | ✓ | Mixed 2.8-bit | 90.45 |
| F. Full ADQ (Ours) | Based on pre-trained model | ✓ | Mixed 3-bit | 91.26 |

The results clearly demonstrate the efficacy of our innovations. First, comparing Config. E (90.45%) with B (90.26%) and C (89.95%) confirms that both our pre-trained model-based initialization and the online adaptive learning mechanism are key to improving performance. More importantly, the superiority of the mixed-precision strategy is doubly validated: under an equivalent bit budget, Config. F (mixed 3-bit, 91.26%) significantly outperforms Config. D (fixed 3-bit, 90.98%). In an efficiency-focused scenario, Config. E achieves highly competitive performance to D with a lower 2.8-bit budget, showcasing an excellent accuracy-efficiency trade-off.

Ultimately, our full method (Config. E) achieves a remarkable >2.24% accuracy improvement over the N2UQ baseline (Config. A, 88.21%), proving the overall superiority of our approach.

4.4 Convergence Analysis

Figure 4.1 shows the training and validation accuracy curves for ADQ-quantized ResNet-18 on ImageNet, with and without knowledge distillation.

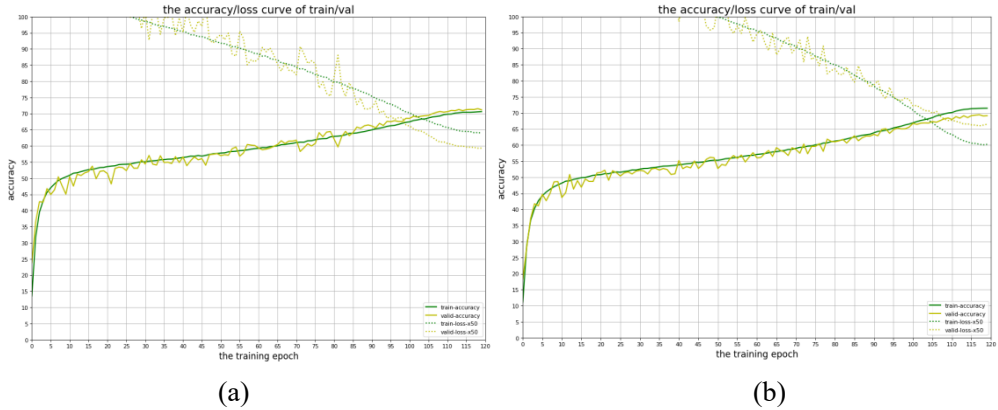

(a) (b)

Figure 4.1 ADQ Quantization Training Curve of ResNet-18 on ImageNet. (a) Top-1 accuracy with knowledge distillation (b) Top-1 accuracy without knowledge distillation

The curves demonstrate that the proposed method achieves stable convergence throughout the training process, without significant fluctuations or performance collapse. This stability can be attributed to the EMA-based online codebook adaptation, which smoothly tracks the weight distribution, and the well-defined commitment loss, which provides a consistent training signal.

## 5 Conclusion and Future Work

In this paper, we proposed ADQ, a novel mixed-precision quantization framework that takes a differentiated and distribution-aware approach to handling weights and activations. Our core contribution is a fully adaptive weight quantization scheme featuring quantile-based initialization, online EMA-based codebook adaptation, and a lightweight, sensitivity-aware mixed-precision strategy. Experiments with ResNet-18 on the ImageNet dataset demonstrate that ADQ achieves state-of-the-art performance, validating the effectiveness of our method.

For future work, we plan to extend the ADQ framework to other architectures, such as Vision Transformers, and explore its application in tasks like object detection. Additionally, we will investigate more sophisticated yet still efficient sensitivity estimation methods to further refine the mixed-precision allocation scheme.

# References

[1] Jacob, B., Kligys, S., Chen, B., Zhu, M., Tang, M., Howard, A., ... & Kalenichenko, D. (2018). Quantization and training of neural networks for efficient integer-arithmetic-only inference. In Proceedings of the IEEE conference on computer vision and pattern recognition (pp. 2704-2713).

[2] Zhou, S., Wu, Y., Ni, Z., Zhou, X., Wen, H., & Zou, Y. (2016). Dorefa-net: Training low bitwidth convolutional neural networks with low bitwidth gradients. arXiv preprint arXiv:1606.06160.

[3] Choi, J., Wang, Z., Venkataramani, S., Chuang, P. I. J., Srinivasan, V., & Gopalakrishnan, K. (2018). Pact: Parameterized clipping activation for quantized neural networks. arXiv preprint arXiv:1805.06085.

[4] Esser, S. K., McKinstry, J. L., Bablani, D., Appuswamy, R., & Modha, D. S. (2020). Learned step-size quantization. In International Conference on Learning Representations.

[5] Zhang, D., Yang, J., Ye, D., & Hua, G. (2018). Lq-nets: Learned quantization for highly accurate and compact deep neural networks. In Proceedings of the European conference on computer vision (ECCV) (pp. 365-382).

[6] Yao, Z., Dong, Z., Zheng, Z., Gholami, A., Yu, J., Tan, E., ... & Keutzer, K. (2020). Hawq-v2: Hessian aware trace-weighted quantization of neural networks. In Proceedings of the IEEE/CVF Conference on Computer Vision and Pattern Recognition (pp. 5355-5364).

[7] Wu, B., Wang, Y., Zhang, P., Tian, Y., Vajda, P., & Keutzer, K. (2018). Mixed precision quantization of convnets via differentiable neural architecture search. arXiv preprint arXiv:1812.00090.

[8] Liu, Z., Cheng, K. T., Huang, D., Shen, Z., & Xing, E. (2022). Nonuniform-to-uniform quantization: Towards accurate quantization via generalized straight-through estimation. In Proceedings of the IEEE/CVF Conference on Computer Vision and Pattern Recognition (pp. 4942-4952).

[9] Y. Bengio, N. Leonard, and A. Courville, "Estimating or propagating gradients through stochastic neurons for conditional computation" arXiv preprint arXiv:1308.3432, 2013.

[10] Y. Bhalgat, J. Lee, M. Nagel, T. Blankevoort, and N. Kwak, "Lsq+: Improving low-bit quantization through learnable offsets and better initialization," in Proc. IEEE Conf. Comput. Vis. Pattern Recog., 2020.

[11] G. Xiao, J. Lin, M. Seznec, H. Wu, J. Demouth, and S. Han, "Smoothquant: Accurate and efficient post-training quantization for large language models," in Proc. Int. Conf. Mach. Learn., 2023.

[12] J. Lin, J. Tang, H. Tang, S. Yang, W.-M. Chen, W.-C. Wang, G. Xiao, X. Dang, C. Gan, and S.

Han, "Awq: Activation-aware weight quantization for llm compression and acceleration," in Proc. Mach. Learn. Syst. Conf., 2024.

[13] Y. Li, R. Gong, X. Tan, Y. Yang, P. Hu, Q. Zhang, F. Yu, W. Wang, and S. Gu, "Brecq: Pushing the limit of post-training quantization by block reconstruction," in Proc. Int. Conf. Learn. Represent., 2021.

[14] Z. Dong, Z. Yao, A. Gholami, M. W. Mahoney, and K. Keutzer, "Hawq: Hessian aware quantization of neural networks with mixed-precision," in Proc. Int. Conf. Comput. Vis., 2019, pp. 293–302.

[15] Z. Dong, Z. Yao, D. Arfeen, A. Gholami, M. W. Mahoney, and K. Keutzer, "Hawq-v2: Hessian aware trace-weighted quantization of neural networks," in Proc. Adv. Neural Inform. Process. Syst., vol. 33, 2020, pp. 18 518–18 529.

[16] Hinton G, Vinyals O, Dean J. Distilling the knowledge in a neural network[J]. arXiv preprint arXiv:1503.02531, 2015.

[17] Van Den Oord A, Vinyals O. Neural discrete representation learning[J]. Advances in neural information processing systems, 2017, 30.

[18] Theis L, Oord A, Bethge M. A note on the evaluation of generative models[J]. arXiv preprint arXiv:1511.01844, 2015.

[19] Molchanov P, Mallya A, Tyree S, et al. Importance estimation for neural network pruning[C]//Proceedings of the IEEE/CVF conference on computer vision and pattern recognition. 2019: 11264-11272.

[20] Deng J, Dong W, Socher R, et al. Imagenet: A large-scale hierarchical image database[C]//2009 IEEE conference on computer vision and pattern recognition. Ieee, 2009: 248-255.

[21] He K, Zhang X, Ren S, et al. Deep residual learning for image recognition[C]//Proceedings of the IEEE conference on computer vision and pattern recognition. 2016: 770-778.

[22] Zhang Z, Shao W, Gu J, et al. Differentiable dynamic quantization with mixed precision and adaptive resolution[C]//International Conference on Machine Learning. PMLR, 2021: 12546-12556.

[23] Yao Z, Dong Z, Zheng Z, et al. Hawq-v3: Dyadic neural network quantization[C]//International Conference on Machine Learning. PMLR, 2021: 11875-11886.

[24] Gong R, Liu X, Jiang S, et al. Differentiable soft quantization: Bridging full-precision and low-bit neural networks[C]//Proceedings of the IEEE/CVF international conference on computer vision. 2019: 4852-4861.

[25] Jung S, Son C, Lee S, et al. Learning to quantize deep networks by optimizing quantization

intervals with task loss[C]//Proceedings of the IEEE/CVF conference on computer vision and pattern recognition. 2019: 4350-4359.

[26] Yamamoto K. Learnable companding quantization for accurate low-bit neural networks[C]//Proceedings of the IEEE/CVF conference on computer vision and pattern recognition. 2021: 5029-5038.

[27] Pei Z, Yao X, Zhao W, et al. Quantization via distillation and contrastive learning[J]. IEEE Transactions on Neural Networks and Learning Systems, 2023.

[28] Kim J, Bhalgat Y, Lee J, et al. Qkd: Quantization-aware knowledge distillation[J]. arXiv preprint arXiv:1911.12491, 2019.